\documentclass[journal]{IEEEtai}

\usepackage[colorlinks,urlcolor=blue,linkcolor=blue,citecolor=blue]{hyperref}

\usepackage{color,array}

\usepackage{graphicx}

\IEEEoverridecommandlockouts
\usepackage{cite}
\usepackage{amsmath,amssymb,amsfonts}
\usepackage{algpseudocode}
\usepackage{algorithm2e}
\usepackage{multirow}
\usepackage{graphicx}
\usepackage{textcomp}
\usepackage{xcolor}
\usepackage[overload]{empheq}
\usepackage{amsmath,amssymb,amsthm,mathtools}
\usepackage{bm}           
\usepackage{bbm}          

\usepackage{subcaption}   

\usepackage{booktabs}     
\usepackage{siunitx}      
\usepackage{multirow}
\usepackage{makecell}

\usepackage{theorem}

\usepackage{microtype}
\newcommand{\method}{FedFusion}
\newcommand{\diven}{DivEn}          
\newcommand{\divenmix}{DivEn-mix}   
\newcommand{\divenc}{DivEn-c}       

\usepackage{subcaption}
\def\BibTeX{{\rm B\kern-.05em{\sc i\kern-.025em b}\kern-.08em
    T\kern-.1667em\lower.7ex\hbox{E}\kern-.125emX}}
\begin{document}

\title{\method: Federated Learning with Diversity- and Cluster-Aware Encoders for Robust Adaptation under Label Scarcity}

\author{\IEEEauthorblockN{Ferdinand Kahenga$^\dag$, Antoine Bagula$^\dag$, Patrick Sello$^\dag$, and Sajal K. Das$\ddag$\\}
$\dag$Department of Computer Science, University of the Western Cape,
Cape Town, South Africa\\
$\ddag$Department of Computer Science,
Missouri University of Science and Technology, Rolla,
Missouri, USA \\
E-mail: 
ferdinandkahenga@esisalama.org,abagula@uwc.ac.za,3919362@myuwc.ac.za, sdas@mst.edu
}

\maketitle

\begin{abstract}
Federated learning in practice must contend with heterogeneous feature spaces, severe non-IID data, and scarce labels across clients. We present \textbf{\method} (\emph{FedFusion}), a federated transfer-learning framework that \emph{unifies} domain adaptation and frugal labelling with \emph{diversity-/cluster-aware encoders} (\diven, \divenmix, \divenc). Labelled \emph{teacher} clients guide \emph{learner} clients via confidence-filtered pseudo-labels and domain-adaptive transfer, while clients maintain \emph{personalised encoders} tailored to local data. To preserve global coherence under heterogeneity, \textbf{\method}  employs \emph{similarity-weighted classifier coupling} (with optional cluster-wise averaging), mitigating dominance by data-rich sites and improving minority-client performance. The frugal-labelling pipeline combines self-/semi-supervised pretext training with selective fine-tuning, reducing annotation demands without sharing raw data. Across tabular and imaging benchmarks under IID, non-IID, and label-scarce regimes, \textbf{\method} consistently outperforms state-of-the-art baselines in accuracy, robustness, and fairness while maintaining comparable communication and computation budgets. These results show that harmonising personalisation, domain adaptation, and label efficiency is an effective recipe for robust federated learning under real-world constraints.
\end{abstract}


\vspace{-0.15in}
\section{Introduction}\label{sec:1}

Federated learning (FL) enables decentralised model training while keeping raw data local, aligning with data minimisation and privacy requirements \cite{kairouz2021}. This paradigm has spurred a broad class of optimisation strategies across academia and industry \cite{FedProx,FedAdam,fedfast}, with particular relevance to privacy-sensitive sectors such as healthcare \cite{youn2023,xu2023,huang2024}. In parallel, transfer learning (TL) can be \emph{combined with} FL—often termed federated transfer learning—to let models trained in one domain generalise to another, thereby reducing reliance on large labelled datasets \cite{long2016}. 

\begin{figure}[h]
\centering
\begin{subfigure}[b]{0.48\linewidth}
    \includegraphics[width=\linewidth]{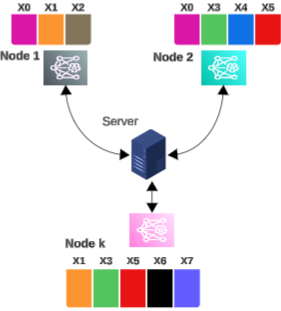}
    \caption{Uneven feature distribution: different features across clients.}
    \label{fig:diven_subfig}
\end{subfigure}
\hfill
\begin{subfigure}[b]{0.48\linewidth}
    \includegraphics[width=\linewidth]{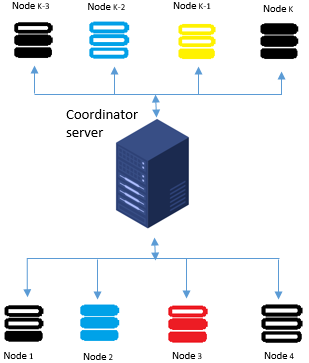}
    \caption{Data distribution: nodes’ data are fully, partially, or not labelled.}
    \label{fig:feddafl_subfig}
\end{subfigure}
\caption{Federated diversity learning: Colour differences indicate variation in feature sets and data distributions.}
\label{fig:problem_scenario}
\end{figure}

Despite this progress, FL degrades in practice when client data are non-IID \cite{li2020}, and global models can underperform under severe heterogeneity \cite{guo2024}. Two patterns are especially common: (i) \emph{feature-space heterogeneity}, where clients expose non-aligned and variably sized feature sets; and (ii) \emph{label scarcity under domain shift}, where clients have fully/partially/unlabelled data from different domains. Figure~\ref{fig:problem_scenario} illustrates two typical scenarios: (a) clients with non-aligned, variably sized feature sets (Fig.~\ref{fig:diven_subfig}); and (b) mixed label availability across domains (Fig.~\ref{fig:feddafl_subfig}). 

In the configuration of Fig.~\ref{fig:feddafl_subfig}, clients may possess fully labelled, partially labelled, or entirely unlabelled data originating from diverse domains, yet all contribute to the same classification task. As an overview, these examples highlight the dual sources of heterogeneity—features and labels—encountered in practice. These settings call for \emph{diversity-aware} personalisation alongside mechanisms that exploit unlabelled data without compromising privacy.

We propose \emph{\method}, a federated transfer-learning framework that integrates \emph{domain adaptation} with \emph{frugal labelling} in a \emph{multi-teacher/multi-learner} setting. \text{\method} addresses: (i) heterogeneous feature partitions in tabular data (common in healthcare, where sources differ despite shared tasks); and (ii) cross-domain variability in images that impairs generalisation. Concretely, \text{\method} combines personalised encoders with \emph{similarity-weighted classifier sharing} (DivEn/DivEn-mix) and a \emph{cluster-aware} variant (DivEn-c) to share structure among similar clients while preserving local specialisation. It further couples a two-step pipeline—self-supervised pretext training followed by confidence-filtered pseudo-label fine-tuning—to leverage unlabelled data under domain shift.

Building on the PRIS line of work, \text{\method} unifies federated learning, transfer learning, and domain adaptation to support both tabular and imaging tasks while reducing dependence on extensive labels \cite{daly2024}. The framework explicitly targets realistic federated deployments (e.g., Gauteng Department of Health) where data are fragmented, sparsely labelled, and domain-diverse.

This article substantially extends our preliminary FedDAFL~\cite{feddafl2024} work by introducing:
\begin{itemize}
  \item \textbf{Personalised diversity/cluster-aware encoders} (\diven/\divenmix/\divenc) with similarity-weighted classifier aggregation to handle heterogeneous feature spaces.
  \item \textbf{Integrated two-step domain-adaptation with frugal labelling} (pretext $\rightarrow$ pseudo-labels), including confidence filtering and encoder-drift control for unlabelled clients.
  \item \textbf{Broader evaluation} across non-IID, label-scarce tabular (heterogeneous features) and imaging (domain shift) benchmarks, with expanded ablations, fairness metrics, and efficiency analyses.
\end{itemize}

The remainder of the paper is organised as follows: Section~II reviews related work; Section~III presents the \textbf{FedFusion} methodology (\diven/\divenmix/\divenc) and the two-step frugal-label domain-adaptation pipeline); Section~IV details the algorithms and training protocols; Section~V reports performance evaluation across tabular feature-distribution settings and imaging domain-adaptation benchmarks; Section~VI provides performance justification; and Section~VII concludes and outlines directions for future work.

\vspace{-0.05in}

\section{Related work} \label{sec:2}
\vspace{-0.05in}
Federated Learning \cite{fedavg} trains global models without sharing local data, thereby preserving user privacy. Despite progress on data heterogeneity and communication constraints, FL still struggles with uneven feature distributions and unlabeled data, limiting applicability in domains such as healthcare. We summarise closely related work in \emph{personalised federated learning} and \emph{domain adaptation within FL}.

\subsection{Personalised Federated Learning}
Personalised FL develops methods that allow clients to learn individualised models while leveraging global knowledge. Zhou et al.~\cite{zhou2025mfed} propose \textit{M-Fed}, a multi-task encoder–decoder framework: clients on distinct tasks maintain local encoders/classifiers; encoders for the same task are aggregated, while a \emph{global} encoder provides shared abstract knowledge with a regulariser controlling its influence during local updates. 
Gain et al.~\cite{gain2025fedfeat} introduce \textit{FedFeat+}, which decomposes models into a feature extractor and a classifier. Clients transmit both components along with differentially private, noise-added features; the server aggregates extractors/classifiers and retrains only the global classifier on noisy labelled features, improving robustness without exposing raw data. 
FedPAC~\cite{xu2023fedpac} employs a personalised strategy using a regulariser of the form $|f(x)-c|$, where $f(x)$ is the local embedding and $c$ is the global class centroid; clients collaborate to compute centroids and weight classifiers, yielding class-aware updates while retaining personalisation.

Complementary to these, correlation-alignment methods (e.g., \cite{coral}) harmonise intermediate representations prior to aggregation by penalising divergence of correlation matrices from a global target, thereby encouraging feature-space consistency across clients.

These directions illustrate an evolution from multi-task collaboration to privacy-preserving model construction and personalised optimisation, via encoder sharing, feature-alignment regularisation, and classifier aggregation. However, most approaches either assume \emph{uniform model architectures} across clients or primarily address \emph{task divergence}. They do not explicitly tackle \emph{unequal and stochastic feature-dimension distributions} across clients in tabular settings, nor do they natively account for mixed labelled/partially labelled/unlabelled regimes. Our approach leverages latent client commonalities even when feature spaces differ, promoting effective collaboration while preserving local structural diversity.

\subsection{Domain Adaptation}
Transfer learning seeks to improve learning in a target task by transferring knowledge from a related source task \cite{tl0}. Within deep transfer learning, two primary cases are common: fine-tuning and domain adaptation. Traditionally, domain adaptation has been studied in centralised settings with joint access to source and target data.

Within FL, \emph{unsupervised federated domain adaptation} methods address decentralised data and label scarcity. FADA~\cite{fada} and FedKA~\cite{fedka} represent distinct strategies. FADA pursues adversarial alignment in settings where both labelled source and unlabelled target data remain inaccessible to each other: it decouples optimisation into domain-specific local feature extractors and a global discriminator, aligning distributions via adversarial training to mitigate domain shift. FedKA, in contrast, focuses on feature-distribution matching to obtain domain-invariant client features for the global model; it employs a federated voting mechanism to derive target pseudo-labels from client consensus, enabling global fine-tuning and reducing negative transfer.

Self-supervised learning (SSL) provides powerful mechanisms to exploit unlabelled data. An early line of work proposes generic self-supervised domain adaptation via rotation prediction as a pretext task \cite{ssl0}, yielding transferable representations that benefit downstream classification.

In summary, while FL addresses privacy in decentralised training, transfer learning and domain adaptation (e.g., FADA) extend FL to unlabelled and distribution-shifted scenarios, and SSL amplifies these benefits by extracting signal from unlabelled data. Nonetheless, existing approaches typically consider single-teacher or limited teacher–learner configurations and assume aligned feature spaces. In contrast, \textbf{FedFusion} targets realistic \emph{multi-teacher/multi-learner} settings with frugal labelling and explicitly handles (i) heterogeneous feature partitions in tabular data and (ii) domain variability in images, providing a unified path toward robust personalisation under non-IID, label-scarce constraints.

\subsection{What gap does \method\ fill?}
Table~\ref{tab:feddafl_vs_method_intro} surfaces three persistent deployment gaps left by preliminary \textsc{FedDAFL}, and how \method\ closes them:

 \textit{Gap 1: Heterogeneous feature spaces in tabular FL.} Real clients rarely share aligned features, making one-size encoders brittle. \textit{\method:} diversity-aware, personalised encoders with \emph{similarity-weighted classifier sharing} (DivEn/DivEn-mix) enable collaboration without forcing identical extractors.

\textit{Gap 2: Mixed label availability under domain shift.} Clients can be fully/partially/entirely unlabeled and sit in different domains. \textit{\method:} integrates the two-step (pretext $\rightarrow$ pseudo-label fine-tuning) domain-adaptation pipeline with the personalised blocks, using confidence filtering and drift control to safely exploit unlabeled data.

\textit{Gap 3: Stability, fairness, and efficiency under non-IID.} Vanilla aggregation can amplify majority clients and destabilise training. \textit{\method:} adds \emph{cluster-aware personalisation} (DivEn-c) and trust-weighted aggregation, and explicitly reports fairness/participation and overheads to make trade-offs transparent.

\noindent Taken together, \method\ is  a capability jump that unifies tabular and imaging settings, handles heterogeneous features and label scarcity, and remains communication-conscious and privacy-preserving in realistic non-IID federated deployments.
\begin{table}[t]
\centering
\caption{Comparison of  \textsc{FedDAFL} vs. \textsc{FedFusion}.}
\label{tab:feddafl_vs_method_intro}
\renewcommand{\arraystretch}{1.08}
\setlength{\tabcolsep}{2pt}
\scriptsize
\begin{tabular}{|p{2.45cm}|p{3.10cm}|p{3.10cm}|}
\hline
\textbf{Aspect} & \textbf{Preliminary \textsc{FedDAFL}} & \textbf{\method\ (FedFusion, this work)} \\
\hline
Problem setting &
MT–ML federated transfer learning under non-IID \& label scarcity (imaging). &
Adds \textit{heterogeneous feature-partition} personalisation (tabular) while unifying tabular \& imaging non-IID, label-scarce regimes. \\
\hline
Encoder design &
Single encoder family aggregated across clients. &
\textbf{DivEn}: diversity-aware (per-client/cluster) encoders; \textbf{DivEn-c} cluster init \& sharing. \\
\hline
Classifier sharing &
No similarity-weighted sharing. &
\textbf{DivEn/DivEn-mix}: similarity-weighted aggregation of classifier heads; optional round-wise reset. \\
\hline
Heterogeneous features (tabular) &
Out of scope. &
\textbf{Supported}: personalised encoders + shared head with similarity regularisation; negative-transfer guard. \\
\hline
Domain adaptation pipeline &
Two-step: pretext for unlabeled/partial, then fine-tune with pseudo-labels. &
Same two-step pipeline \textit{integrated} with DivEn personalisation; optional consistency regularisation. \\
\hline
Benchmarks &
Digits-Five; Chest X-rays. &
\textbf{Plus} tabular feature-distribution suites (Obesity, Heart Disease, Lifestyle; 8/10/12/14 features). \\
\hline
Analyses &
Limited ablations. &
\textbf{Expanded}: DivEn vs DivEn-mix vs DivEn-c, client-level, feature-count sweeps, threshold sensitivity. \\
\hline
Fairness/efficiency &
Basic timing. &
\textbf{Added} fairness/balanced-participation reporting; efficiency tables and overhead discussion. \\
\hline
\end{tabular}
\end{table}


\section{Methods (\textbf{FedFusion})} \label{sec:3}

\subsection{Diversity-Aware Encoding under Heterogeneous Feature Partitions \textbf{(DivEn)}}

Traditional FL often assumes homogeneous feature sets across clients, a premise that rarely holds in real-world decentralised environments. This leads to the challenge of learning personalised models when local datasets exhibit significant feature heterogeneity. Despite these differences, useful commonalities in distributions or task structure can be leveraged.

\subsection*{Problem Formulation and Methods}
We model each client’s personalised model $\mathcal{M}_i$ as a composition of an \emph{encoder} $\mathcal{E}_i$ and a \emph{classifier} $\mathcal{C}_i$, i.e., $\mathcal{M}_i = \mathcal{C}_i \circ \mathcal{E}_i$. Let $D^i = \{(x_k^i, y_k^i)\}_{k=1}^{N_i}$ denote client $i$’s local dataset, and let $S_i \subset F_{\text{all}}$ be the feature subset used by client $i$, where $x_k^i$, $y_k^i$, $N_i$, and $F_{\text{all}}$ represent, respectively, the $k$-th input sample of client $i$, its corresponding label, the total number of samples held by client $i$, and the global set of all possible features.
 We allow $S_i\neq S_j$ for $i\neq j$ (feature heterogeneity), while assuming a non-empty overlap $\bigcap_{i=1}^K S_i \neq \emptyset$, where $K$ is the total number of clients participating in the FL process.

\textbf{DivEn} addresses this via:
\begin{enumerate}
\item \textbf{Personalised encoder adaptation.} Each client $i$ selects its encoder $\mathcal{E}_i$ (e.g., via Bayesian hyperparameter optimisation) tuned to $D^i$ and $S_i$. The classifier head shares a fixed architecture to promote a common aggregation interface.
\item \textbf{Similarity-weighted classifier regularisation/aggregation.} While encoders are personalised, clients exchange classifier knowledge. Let $\theta_{\mathcal{C}_i}$ denote client $i$’s classifier parameters, and $z_i=\mathcal{E}_i(x,\theta_{\mathcal{E}_i})$ its latent. The server computes a cosine-similarity matrix on latents and builds a \emph{client-specific global classifier} $\theta_{\mathcal{C}_{G_i}}$ via softmax-weighted averaging of peers’ classifiers.
\end{enumerate}

For \textbf{DivEn}, the global classifier acts as a \emph{regulariser} during local training. At round $r$, client $i$ minimises
\[
L_i \;=\; \mathrm{CE}(y,\hat y)\;+\;\lambda \sum_{l}\left\|\theta^{(r)}_{\mathcal{C}_i,l}-\theta^{(r-1)}_{\mathcal{C}_{G_i},l}\right\|_2^2,
\]
where $\hat y=\mathcal{C}_i(z_i;\theta_{\mathcal{C}_i})$ and $\lambda>0$. In \textbf{DivEn-mix}, in addition to this regularisation, the local classifier is \emph{reset} each round to the aggregated one, $\theta_{\mathcal{C}_i}^{(r)}\!\leftarrow\!\theta_{\mathcal{C}_{G_i}}^{(r)}$.

The overall objective is
\[
\min_{\{\theta_{\mathcal{E}_i},\theta_{\mathcal{C}_i}\}_{i=1}^{K}} \;\sum_{i=1}^{K} L_i,
\]
thereby coupling personalised encoder learning with similarity-aware classifier sharing.

\subsubsection{DivEn improvement using clustering (DivEn-c)}
\textbf{DivEn-c} augments DivEn with a feature-space clustering pre-process. Clients are grouped into clusters $\mathcal{C}_{\text{final}}{=}\{C_1,\dots,C_J\}$ based on feature-set similarity (e.g., Jaccard over $S_i$ vectors). For each cluster $C_j$, we identify strictly overlapping features $O_j$ common to all members. Clustering proceeds via an initial KMeans (with $J$ chosen by silhouette/elbow) followed by recursive refinement until intra-cluster similarity exceeds a threshold (e.g., $80\%$), otherwise split to singletons.

Using the resulting clusters, we first \emph{intra-cluster} aggregate encoders/classifiers (size-weighted), then proceed with similarity-weighted classifier sharing as in DivEn.

\subsection{Domain Adaptation (\textbf{FedFusion})}
\subsubsection{Problem statement}
Consider $K$ clients learning collaboratively under privacy constraints. Client $i$ owns private data $D^i=D_{\mathrm{lb}}^{i}\cup D_{\mathrm{ul}}^{i}$ with labelled and unlabelled parts, respectively. Data are non-identically distributed: $\mathrm{dist}(D^{a})\neq \mathrm{dist}(D^{b})$. We categorise clients as: (1) fully labelled ($|D_{\mathrm{ul}}^i|{=}0$), (2) partially labelled ($|D_{\mathrm{ul}}^i|{>}0$, $|D_{\mathrm{lb}}^i|{>}0$), and (3) fully unlabelled ($|D_{\mathrm{lb}}^i|{=}0$).

Let the encoder $E$ map an image to a $d$-dimensional feature, $M$ be the $k$-class main-task head, and $P$ the $v$-class pretext head:
\begin{equation} \label{eq:pretrain}
E:\,\mathbb{R}^{w\times h \times 3} \rightarrow \mathbb{R}^{d},\qquad
M:\,\mathbb{R}^{d} \rightarrow \mathbb{R}^{k},\qquad
P:\,\mathbb{R}^{d} \rightarrow \mathbb{R}^{v}.
\end{equation}
At each round, the server aggregates $\theta_e$ (encoder params) and broadcasts back to clients. The global objective couples task and pretext:
\begin{equation} \label{eq:pretext_loss}
\underset{\theta_{e},\theta_{m},\theta_{p}}{\min}\;\; \mathcal{L}^{\mathrm{task}}(\theta_{e},\theta_{m}) \;+\; \lambda\, \mathcal{L}^{\mathrm{pre}}(\theta_{e},\theta_{p}),
\end{equation}
with per-client losses
\begin{equation}\label{eq:pretext_loss_m}
\mathcal{L}^{\mathrm{task}}_m(\theta_{e},\theta_{m})=\frac{1}{B} \sum_{b=1}^{B} \mathcal{H}\!\left(y_b,\;M(E(x_b))\right),
\end{equation}
\begin{equation}\label{eq:pretext_loss_p}
\mathcal{L}^{\mathrm{pre}}_m(\theta_{e},\theta_{p})=\frac{1}{B} \sum_{b=1}^{B} \mathcal{H}\!\left(\tilde{y}_b,\;P(E(\mathrm{Rot}(u_b)))\right),
\end{equation}
where $\mathcal{H}$ is cross-entropy and $\tilde{y}_b\!\in\!\{0,\dots,v{-}1\}$ is the pretext label (e.g., rotation).

\subsubsection{Self-learning (two-step) in FedFusion}
We employ two steps: (i) \emph{self-supervised pretext} to obtain domain-invariant features, (ii) \emph{task fine-tuning} with frugal labels and confidence-filtered pseudo-labels. After each local phase, encoders are aggregated.

\begin{figure}[h]
\centering
\begin{subfigure}[h]{0.23\textwidth}
\centering
\includegraphics[width=\textwidth]{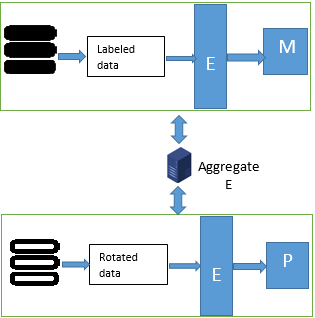}
\caption{Step 1: Pretext}\label{fig:step1}
\end{subfigure}\quad
\begin{subfigure}[h]{0.23\textwidth}
\centering
\includegraphics[width=\textwidth]{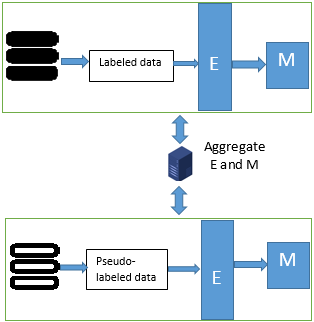}
\caption{Step 2: Fine-tune}\label{fig:step2}
\end{subfigure}
\caption{FedFusion two-step self-learning pipeline.}
\label{fig:steps}
\end{figure}

\paragraph{Step 1 (Pretraining).}
Clients with labels train $E{+}M$ using Eq.~\eqref{eq:pretext_loss_m}; fully unlabelled clients train $E{+}P$ using Eq.~\eqref{eq:pretext_loss_p}. The server aggregates $\theta_e$ each round via size-weighted averaging.

\paragraph{Step 2 (Fine-tuning).}
We couple supervised and semi-/self-supervised learning. With weak/strong augmentations $\Phi,\Psi$ and encoder $f(\cdot){=}E(\cdot)$, a consistency term can be added:
\begin{equation}\label{eq:consistency_loss}
\frac{1}{n}\sum_{i=1}^{n} \left\| f(\Phi(u_i)) - f(\Psi(u_i)) \right\|_2^{2}.
\end{equation}
Extending FixMatch-style training~\cite{fix_match}, define
\begin{equation}\label{e:l1}
\mathcal{L}_1=\frac{1}{B}\sum_{b=1}^{B}\mathcal{H}\!\left(y_b,\;M\!\left(E(\Phi(x_b))\right)\right),
\end{equation}
\begin{equation}\label{e:l2}
\mathcal{L}_2=\frac{1}{B}\sum_{b=1}^{B}\mathbf{1}\!\left(\max(q_b)\ge \tau\right)\,
\mathcal{H}\!\left(\tilde{q}_b,\;M\!\left(E(\Psi(u_b))\right)\right),
\end{equation}

with $q_b = M(E(\Phi(u_b)))$, $\tilde{q}_b = \arg\max(q_b)$, and threshold $\tau \in (0,1)$. Loss by client type:  
\text{fully labelled:} $\mathcal{L}_m = \mathcal{L}_1$; \quad  
\text{partially labelled:} $\mathcal{L}_m = \mathcal{L}_1 + \alpha \mathcal{L}_2$; \quad  
\text{fully unlabelled:} $\mathcal{L}_m = \mathcal{L}_2$.

To limit encoder drift from noisy pseudo-labels, fully unlabelled clients update \emph{only} $\theta_m$ in Step 2 (design choice that can be adapted); labelled/partially labelled clients update both $\theta_e$ and $\theta_m$.
\section{Algorithms and Training Protocols (\textbf{FedFusion})} \label{sec:4}

\subsection{DivEn (Similarity-Weighted Classifier Sharing)}

\begin{algorithm}[h!]
\caption{DivEn: Diversity-Aware Encoding (similarity-weighted classifier sharing)}
\label{alg:divEn}
\KwIn{$N$ clients with datasets $(X_i,y_i)$; rounds $R$; local epochs $E_{\mathrm{init}},E_{\mathrm{low}}$; regularisation $\lambda$; softmax temperature $\tau$; latent dim $\ell$; classifier arch. $\mathcal{A}$}
\KwOut{Personalised models $\mathcal{M}_1,\dots,\mathcal{M}_N$}
\For{each client $i\in\{1,\dots,N\}$}{
  Select $\mathcal{E}_i^\star$ (e.g., Bayesian search) using fixed classifier arch. $\mathcal{A}$; set $\theta_{\mathcal{C}_{G_i}}^{(0)}\leftarrow\text{None}$\;
}
\For{$r\leftarrow 1$ \KwTo $R-1$}{
  Set $E_r \leftarrow E_{\mathrm{init}}$ if $r{=}1$, else $E_{\mathrm{low}}$\;
  \tcp{Local update}
  \For{each client $i$}{
    Receive $\theta_{\mathcal{C}_{G_i}}^{(r-1)}$; train for $E_r$ epochs with
    \[
      \mathcal{L}_i \!=\! \mathrm{CE}(y,\hat y) + \lambda \sum_{l}\!\left\|\theta^{(r)}_{\mathcal{C}_i,l}\!-\!\theta^{(r-1)}_{\mathcal{C}_{G_i},l}\right\|_2^2;
    \]
    update $(\theta^{(r)}_{\mathcal{E}_i},\theta^{(r)}_{\mathcal{C}_i})$; record $\mathrm{acc}_i^{(r)}$;\\
    if $r{=}1$: store $\texttt{threshold\_acc}_i\!\leftarrow\!\mathrm{acc}_i^{(1)}$, $\texttt{threshold\_params}_i\!\leftarrow\!\texttt{get\_weights}(\mathcal{M}_i)$\;
  }
  \tcp{Latent similarity and weights}
  \For{each client $i$}{
    $z_i^{(r)}\leftarrow \mathcal{E}_i^{(r)}(X_i)$\;
    \For{each client $j$}{ $s_{ij}^{(r)}\leftarrow \texttt{cos\_sim}(z_i^{(r)},z_j^{(r)})$ }
    $\alpha_{ij}^{(r)} \leftarrow \exp(s_{ij}^{(r)}/\tau)\Big/\sum_{k=1}^{N}\exp(s_{ik}^{(r)}/\tau)$\;
  }
  \tcp{Classifier aggregation (DivEn); optional reset (DivEn-mix)}
  \For{each client $i$}{
    $\theta_{\mathcal{C}_{G_i}}^{(r)} \leftarrow \sum_{j=1}^{N} \alpha_{ij}^{(r)}\, \theta_{\mathcal{C}_j}^{(r)}$\;
    \If{\texttt{DivEn-mix}}{ $\theta_{\mathcal{C}_i}^{(r)} \leftarrow \theta_{\mathcal{C}_{G_i}}^{(r)}$ }
  }
}
\tcp{Negative-transfer guard}
\For{each client $i$}{
  \If{$\mathrm{acc}_i^{(R-1)} < \texttt{threshold\_acc}_i$}{
    Revert to $\texttt{threshold\_params}_i$; brief local retraining; accept only if accuracy improves\;
  }
}
\Return{$\{\mathcal{M}_i\}_{i=1}^{N}$}
\end{algorithm}

The algorithm~\ref{alg:divEn} begins by allowing each client to train its model extensively ($E_{init}$ epochs) during the initial round. Upon completion, each client evaluates and stores two items: the performance metric, denoted as \textit{threshold\_acc}, and the model parameters, denoted as \textit{threshold\_params}. Each client then transmits its local classifier parameters ($\mathbf{\theta}_{\mathcal{C}_i}$) along with its latent representation, $z_i=\mathcal{E}_i(x,\theta_{\mathcal{E}_i})$, to the server.

The server assesses inter-client proximity by comparing latent representations across clients. Specifically, it computes the cosine distances between latent spaces to derive a client similarity matrix. Based on this matrix, the server constructs each client's global classifier ($\theta_{\mathcal{C}_{G_i}} $) using a weighted aggregation of peer classifiers.

Subsequent training rounds involve only a few local epochs per client ($E_{low}$ epochs), thereby updating  each client's parameters ($\mathbf{\theta}_{\mathcal{E}_i},\theta_{\mathcal{C}_{i}}$). After the required number of rounds, a quick performance validation step is performed to avoid negative transfer: each client locally verifies whether its current performance surpasses the stored threshold. If not, it replaces its model parameters with the previously saved ones (\textit{threshold\_params}) and performs brief local retraining.

\subsection{DivEn-c (Client Feature-Based Clustering)}
Following the clustering process described in Algorithm~\ref{alg:div-c-clusteringphase}, the original \textit{DivEn} algorithm~\ref{alg:divEn} is refined into Algorithm~\ref{alg:divEn_clustering} to support optimal encoder discovery within each of the $K$ clusters. Since clients within a given cluster share the same feature subsets, the strategy involves randomly selecting one representative client per cluster to perform encoder optimisation on behalf of the group. 

Model aggregation is then conducted in two stages: first, local models within each cluster are aggregated to form cluster-specific encoders and classifiers; second, the refined algorithm proceeds using these aggregated cluster models as inputs to the subsequent steps.
\begin{algorithm}[h!]
\caption{Feature-space-based Clustering (client feature subsets)}
\label{alg:div-c-clusteringphase}
\KwIn{$\mathcal{F}_{\text{all}}$ (all features), $\{S_i\}$ (client feature subsets), threshold $\texttt{min\_sim}$}
\KwOut{$\mathcal{C}_{\text{final}}$ (final clusters), $\{O_k\}$ (overlapping features per cluster)}
Build binary matrix $\mathcal{W}$ from $\{S_i\}$; run KMeans (choose $K$ by silhouette/elbow) for initial groups\;
Recursively refine clusters: if size is large and similarity $<\texttt{min\_sim}$, re-cluster; else stop\;
For each final cluster $C_k$, compute $O_k=\bigcap_{i\in C_k}S_i$ and keep/sort only overlapping features\;
\end{algorithm}

\begin{algorithm}[h!]
\caption{DivEn-c: Diversity Encoding with Client Feature-Based Clustering}
\label{alg:divEn_clustering}
\KwIn{Clusters $\{C_k\}_{k=1}^K$; datasets $(X_i,y_i)$ where $N_i = |X_i|$; rounds $R$; local epochs $E$; regularisation $\lambda$; temperature $\tau$; LR $\eta$; latent dim $\ell$; classifier arch. $\mathcal{A}$}
\KwOut{Personalised models $\{\mathcal{M}_i\}$}
\textbf{Cluster init:} For each $C_k$, pick a representative $i^\star\!\in\!C_k$; optimise $\mathcal{E}_{i^\star}$ (Bayesian search) using fixed classifier arch. $\mathcal{A}$; set $\mathcal{E}_i\!\leftarrow\!\mathcal{E}_{i^\star}$ for $i\in C_k$\;
\For{$r\leftarrow 1$ \KwTo $R-1$}{
  \tcp{Local training as in Alg.~\ref{alg:divEn}}
  \tcp{Intra-cluster aggregation (size-weighted)}
  \For{each cluster $C_k$}{
    $T_k \leftarrow \sum_{j\in C_k} N_j$\;
    $\theta_{\mathcal{E}_{G,k}}^{(r)} \leftarrow \sum_{j\in C_k}\frac{N_j}{T_k}\theta_{\mathcal{E}_j}^{(r)}$;\quad
    $\theta_{\mathcal{C}_{G,k}}^{(r)} \leftarrow \sum_{j\in C_k}\frac{N_j}{T_k}\theta_{\mathcal{C}_j}^{(r)}$\;
    \For{each $i\in C_k$}{ $\theta_{\mathcal{E}_i}^{(r)}\!\leftarrow\!\theta_{\mathcal{E}_{G,k}}^{(r)}$;\; $\theta_{\mathcal{C}_i}^{(r)}\!\leftarrow\!\theta_{\mathcal{C}_{G,k}}^{(r)}$ }
  }
  \tcp{Similarity-weighted classifier sharing (DivEn)}
  Compute $\{\alpha_{ij}^{(r)}\}$ as in Alg.~\ref{alg:divEn}; update $\theta_{\mathcal{C}_{G_i}}^{(r)} = \sum_j \alpha_{ij}^{(r)}\theta_{\mathcal{C}_j}^{(r)}$; optionally reset $\theta_{\mathcal{C}_i}^{(r)}$\;
}
\Return{$\{\mathcal{M}_i\}$}
\end{algorithm}

\subsection{FedFusion Two-Step Self-Learning}

The FedFusion Two-Step Self-Learning algorithm unfolds in the following two steps:

\vspace{0.5em}
\noindent\textbf{Step 1:} As presented in Algorithm~\ref{alg:main_algo}, the coordinator server sends the initial encoder parameters $\theta_e(0)$ to each client. Based on its data status, each client constructs either an $E+P$ model with parameters $\theta=(\theta_e,\theta_p)$ or an $E+M$ model with parameters $\theta=(\theta_e,\theta_m)$, and trains it locally. Fully and partially labelled clients train $E+M$ using Eq.~(\ref{eq:pretext_loss_m}), while fully unlabelled clients train $E+P$ using Eq.~(\ref{eq:pretext_loss_p}), as detailed in Algorithm~\ref{alg:update_client_encoder}. After local training, only the encoder parameters $\theta_e$ are sent back to the server, which aggregates the updates to produce a generalized encoder. This process is repeated across multiple rounds.

\vspace{0.5em}
\noindent\textbf{Step 2:} The coordinator server requests data status from all connected clients, each of which responds accordingly. Using the final encoder parameters $\theta_e(T)$ from Step 1, the server constructs the task model $E+M$ with parameters $W=(\theta_e,\theta_m)$ and distributes it to the clients. Each client trains the received model based on its data status: fully labelled clients use loss $\mathcal{L}_1$ as in Eq.~(\ref{e:l1}); partially labelled clients use both $\mathcal{L}_1$ and $\mathcal{L}_2$ as in Eq.~(\ref{e:l1}) and Eq.~(\ref{e:l2}); and fully unlabelled clients use only $\mathcal{L}_2$ as in Eq.~(\ref{e:l2}). Pseudo-labels for unlabelled clients are selected based on predictions from the global model with confidence exceeding threshold $\tau$. This step corresponds to training Step 2 in Algorithm~\ref{alg:main_algo} and is further detailed in Algorithm~\ref{alg:client_update_step2}.

\begin{algorithm}[h!]
\caption{FedFusion (Two-Step Self-Learning with Domain Adaptation)}
\label{alg:main_algo}
\SetKw{KwInit}{initialize}
\KwInit $\theta_e(0)$; total rounds $T$; client set $\mathcal{N}$; LR $\eta$; sample sizes $\{n_k\}$\;
\textbf{Status map:} $status(k)\in\{1{:}$ fully labelled$,$ 2{:}$ partially labelled$,$ 3{:}$ fully unlabelled$\}$\;

\tcp{Step 1: Pretext / mixed training}
\For{$t\leftarrow 1$ \KwTo $T$}{
  \ForEach{$k\in\mathcal{N}$}{
    $\theta_{e_k}(t)\leftarrow \texttt{UpdateClientEncoder}\big(\theta_e(t{-}1),\,k\big)$\;
  }
  $\theta_e(t)\leftarrow \sum_{k\in\mathcal{N}} \frac{n_k}{\sum_{j\in\mathcal{N}} n_j}\,\theta_{e_k}(t)$\;
}
$W_0 \leftarrow (\theta_e(T),\,\theta_m)$\;

\tcp{Step 2: Fine-tuning (task)}
\For{$t\leftarrow 1$ \KwTo $T$}{
  \ForEach{$k\in\mathcal{N}$}{
    $W_k(t)\leftarrow \texttt{UpdateClient}\big(W(t{-}1),\,k\big)$\;
  }
  $W(t)\leftarrow \sum_{k\in\mathcal{N}} \frac{n_k}{\sum_{j\in\mathcal{N}} n_j}\, W_k(t)$\;
}
\end{algorithm}

\begin{algorithm}[h!]
\caption{UpdateClientEncoder — Step 1}
\label{alg:update_client_encoder}
\SetKwFunction{UpdE}{UpdateClientEncoder}
\UpdE{$\theta_e(t{-}1),\,k$}{\\
  $\theta_e \leftarrow \theta_e(t{-}1)$\;
  \If{$status(k)\in\{1,2\}$}{%
    \tcp{label present: task loss}
    $\theta=(\theta_e,\theta_m)\leftarrow \theta - \eta \nabla \mathcal{L}^{\mathrm{task}}_k(\theta_e,\theta_m)$\;
  }\Else{%
    \tcp{no labels: pretext loss}
    $\theta=(\theta_e,\theta_p)\leftarrow \theta - \eta \nabla \mathcal{L}^{\mathrm{pre}}_k(\theta_e,\theta_p)$\;
  }
  \KwRet $\theta_e$\;
}
\end{algorithm}

\begin{algorithm}[h!]
\caption{UpdateClient — Step 2}
\label{alg:client_update_step2}
\SetKwFunction{Upd}{UpdateClient}
\Upd{$W(t{-}1),\,k$}{%
  \uIf{$status(k){=}1$}{%
    $W_k(t) \leftarrow W(t{-}1) - \eta \nabla \mathcal{L}_1$\;  \tcp{$\mathcal{L}_1$ in Eq.~\eqref{e:l1}}
  }\uElseIf{$status(k){=}2$}{%
    $W_k(t) \leftarrow W(t{-}1) - \eta \nabla \big(\mathcal{L}_1 + \mathcal{L}_2\big)$\; \tcp{$\mathcal{L}_2$ in Eq.~\eqref{e:l2}}
  }\Else{%
    \tcp{fully unlabelled: pseudo-labels (confidence $\tau$)}
    $W_k(t) \leftarrow W(t{-}1) - \eta \nabla \mathcal{L}_2$\;
    \tcp{design choice: update $\theta_m$; keep $\theta_e$ fixed to reduce drift}
  }
  \KwRet $W_k(t)$\;
}
\end{algorithm}

\subsection{Convergence Analysis of \textbf{FedFusion}}
\label{sec:conv-fedf}

\noindent\textbf{What we want to show.}
FedFusion performs local SGD at each client and adds a \emph{similarity pull} on the classifier head so that clients with similar data gently align. We show that (i) the global objective decreases on average each round, (ii) the method reaches a first–order stationary point at the usual $O(1/\sqrt{T})$ rate, up to a \emph{small, interpretable residual} caused by stochasticity, heterogeneity, similarity estimation, clustering, and pseudo-labels; and (iii) under a PL condition it converges linearly to such a neighbourhood.

\paragraph{Setup (objective and notation).}
Client $i$ keeps encoder/classifier parameters $(\theta_{E_i},\theta_{C_i})$ and minimises
\[
f_i(\theta_{E_i},\theta_{C_i})
=\mathbb{E}_{(x,y)\sim\mathcal{D}_i}\!\big[\ell(\theta_{E_i},\theta_{C_i};x,y)\big]
+\lambda\Big\|\theta_{C_i}-\sum_{j=1}^M\alpha_{ij}\theta_{C_j}\Big\|_2^2,
\]
Since the term $\sum_{j=1}^M \alpha_{ij} \theta_{C_j}$ is computed on the server side and transmitted to client $i$, the client does not require direct access to the individual parameters of other clients.

where $A=[\alpha_{ij}]$ is a row-stochastic similarity matrix built from client features. The global objective is
\[
F(\Theta)=\sum_{i=1}^M p_i\, f_i(\theta_{E_i},\theta_{C_i}),\qquad
p_i=\frac{|\mathcal{D}_i|}{\sum_j|\mathcal{D}_j|}.
\]
(Optional) \textbf{DivEn-mix}: hard-reset $\theta_{C_i}$ to the similarity average each round.
(Optional) \textbf{DivEn-c}: do a size-weighted cluster average inside clusters $\{C_k\}$, then apply the similarity step.

\paragraph{One round of FedFusion (mechanics).}
Each selected client runs $E$ steps of local SGD with stepsize $\eta$:
\[
\theta_{E_i}^{t+}=\theta_{E_i}^t-\eta\,\widehat{\nabla}_{\theta_{E_i}} f_i,\qquad
\theta_{C_i}^{t+}=\theta_{C_i}^t-\eta\,\widehat{\nabla}_{\theta_{C_i}} f_i,
\]
and the gradient of the similarity pull is
\[
\widehat{\nabla}_{\theta_{C_i}}\!\left[\lambda\left\|\theta_{C_i}-\textstyle\sum_j\alpha_{ij}\theta_{C_j}^t\right\|_2^2\right]
=2\lambda\left(\theta_{C_i}^t-\textstyle\sum_j\alpha_{ij}\theta_{C_j}^t\right).
\]
The server averages per block with weights $p_i$ and broadcasts.

\paragraph{Assumptions (plain-English).}
\begin{itemize}
\item \textbf{A1 Smoothness:} each $f_i$ has $L$-Lipschitz gradients (standard in SGD).
\item \textbf{A2 Bounded stochasticity:} variance of noisy gradients $\le\sigma^2$.
\item \textbf{A3 Client drift:} gradients of $f_i$ need not match the global gradient; their average deviation is bounded by $B^2$.
\item \textbf{A4 Similarity quality:} the graph $A$ is reasonably mixed (gap $\gamma>0$) and similarity estimates have error $\varepsilon_{\rm sim}$.
\item \textbf{A5 Cluster residual (DivEn-c):} after clustering, within-cluster mismatch is $\Delta_c^2$.
\item \textbf{A6 Pseudo-label bias:} in the semi/self-supervised phase the filtering induces a bounded bias $\varepsilon_{\rm pl}^2$.
\end{itemize}
For compactness define
\[
R_{\rm sim}\triangleq \lambda^2\!\big(\gamma^{-2}\varepsilon_{\rm sim}^2+\Delta_c^2\big),\qquad
\Xi\triangleq \tfrac{L}{2}\sigma^2+B^2+R_{\rm sim}+\varepsilon_{\rm pl}^2.
\]

\paragraph{Step 1 — one-round descent.}
Using smoothness (A1) and the update above, one communication round satisfies
\[
\mathbb{E}\,F(\Theta_{t+1})
\;\le\;
\mathbb{E}\,F(\Theta_t)
-\frac{\eta}{2}\,\mathbb{E}\|\nabla F(\Theta_t)\|^2
+\eta\,\Xi,
\quad \text{for }\eta\le 1/L.
\]
\emph{Interpretation.} The expected global loss drops by a term proportional to the squared gradient, up to an additive “noise floor” $\eta\,\Xi$ that bundles the five error sources.

\paragraph{Step 2 — summing over rounds (non-convex case).}
Telescoping the descent over $T$ rounds yields
\begin{align*}
\min_{0 \le t < T} \mathbb{E} \left\| \nabla F(\Theta_t) \right\|^2
\;\le\;&\ 
\frac{2\left(F(\Theta_0) - F^\star\right)}{\eta T} \\
&+ C_1 \eta \sigma^2 
+ C_2 B^2 
+ C_3 R_{\mathrm{sim}} 
+ C_4 \varepsilon_{\mathrm{pl}}^2
\end{align*}

with constants $C_1,\ldots,C_4$ independent of $T$.
\emph{Takeaway.} With a constant $\eta\!\le\!1/(2L)$ we reach a first-order stationary point at rate $O(1/T)$ in the leading term, up to a residual determined by data noise ($\sigma^2$), heterogeneity ($B^2$), similarity/cluster quality ($R_{\rm sim}$), and pseudo-labels ($\varepsilon_{\rm pl}^2$).

\paragraph{Step 3 — diminishing stepsize.}
With $\eta_t=\eta_0/\sqrt{t+1}$,

\begin{align}
\min_{0\le t<T}\mathbb{E}\|\nabla F(\Theta_t)\|^2
&= \mathcal{O}\!\left(T^{-1/2}\right) \nonumber\\
&\quad + \mathcal{O}\!\left(\sigma^2 T^{-1/2}\right) \nonumber\\
&\quad + \mathcal{O}\!\left(B^2 + R_{\rm sim} + \varepsilon_{\rm pl}^2\right).
\end{align}

\emph{Takeaway.} We still get the standard $1/\sqrt{T}$ non-convex rate; the asymptotic neighbourhood is smaller when the graph mixes well (large $\gamma$), similarities are accurate (small $\varepsilon_{\rm sim}$), clusters are tight (small $\Delta_c$), and pseudo-labels are clean (small $\varepsilon_{\rm pl}$).

\paragraph{PL (Polyak–Łojasiewicz) case.}
If $\tfrac{1}{2}\|\nabla F(\Theta)\|^2\ge \mu\big(F(\Theta)-F^\star\big)$ and $\eta\le 1/L$,
\[
\mathbb{E}\!\left[F(\Theta_{t+1})-F^\star\right]
\le (1-\eta\mu)\,\mathbb{E}\!\left[F(\Theta_t)-F^\star\right]+\eta\,\Xi/\mu,
\]
i.e., linear convergence to a neighbourhood whose radius scales with $\sigma^2,B^2,R_{\rm sim},\varepsilon_{\rm pl}^2$.

\paragraph{How the design choices affect the bound (cheat-sheet).}
\begin{center}
\begin{tabular}{|p{2.0cm}|p{1.2cm}|p{4.0cm}|}
\hline
\textbf{Knob} & \textbf{Term} & \textbf{Effect} \\
\hline
Similarity graph quality (mixing gap $\gamma$) & $R_{\rm sim}$ & Better mixing (larger $\gamma$) $\Rightarrow$ smaller $R_{\rm sim}$.\\
\hline
Similarity accuracy ($\varepsilon_{\rm sim}$) & $R_{\rm sim}$ & More reliable similarities $\Rightarrow$ smaller residual.\\
\hline
Clustering tightness ($\Delta_c$) & $R_{\rm sim}$ & Tighter clusters (DivEn-c) $\Rightarrow$ smaller residual.\\
\hline
Pseudo-label threshold $\tau$ & $\varepsilon_{\rm pl}$ & Higher $\tau$ filters noise $\Rightarrow$ smaller bias but fewer unlabeled samples.\\
\hline
Regulariser weight $\lambda$ & $R_{\rm sim}$ & Larger $\lambda$ enforces agreement but increases $R_{\rm sim}$ quadratically; tune for stability vs.\ residual size.\\
\hline
DivEn-mix reset & (implicit) & Reduces classifier disagreement (shrinks similarity-related constants), may slow per-client personalisation.\\
\hline
\end{tabular}
\end{center}

\paragraph{Practical recipe.}
Use a modest $\eta\!\le\!1/(2L)$ or a decaying schedule; build a well-connected similarity graph (avoid near-disconnected clients); prefer DivEn-c when clients form natural clusters; set a conservative pseudo-label threshold at early rounds and relax it as the encoder improves; and ablate $\lambda,\tau$ and graph construction to verify the predicted trends.

\medskip\noindent
\emph{In short:} FedFusion inherits the standard non-convex FL rates and converges \emph{up to a small, controllable neighbourhood} whose size is governed by five intuitive quantities—noise, heterogeneity, similarity quality, clustering quality, and pseudo-label bias.

\section{Performance Evaluation (\textbf{FedFusion})} \label{sec:4}
\subsection{Feature distribution}
To evaluate the performance of personalised models created after classifier parameter aggregation and/or regularisation, we devised several feature distribution scenarios, which we termed 8-features, 10-features, 12-features, and 14-features. Each of these scenarios highlights the maximum number of features a client can possess. We conducted tests on three datasets: the Obesity dataset~\cite{obesity} (with a total of 16 features), the Heart Diseases dataset (maximum 17 features), and the Life Style dataset~\cite{lifeexpect} (maximum 21 features). The first two datasets were utilised for the classification task, while the last dataset was employed for a linear regression task, using MAE (Mean Absolute Error) as the evaluation metric. 

Our methods were compared against several baselines: (1) a scenario where each client trains its model independently (named Single), (2) an alignment-based method using Coral (named align-corr), (3) a technique based on autoencoders (named Auto-enc), and (4) a method involving full aggregation of classifiers (named class-agg).

\begin{figure}[h]
\centering
\includegraphics[width=\linewidth]{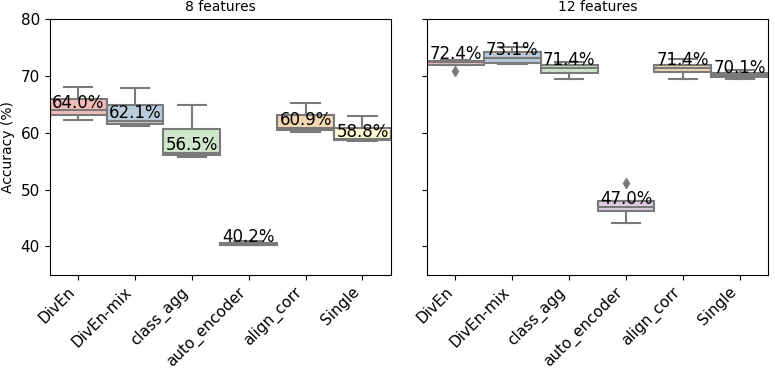}
\caption{Methods comparison across different 8 and 12 feature distribution settings.}
\label{fig:8_12_features}
\end{figure}

Fig.~\ref{fig:8_12_features} shows the results of methods comparison on the Obesity dataset, while Tab.~\ref{tab:side_by_side_tables} presents the values obtained for classification and regression on the 8-features and 12-features scenarios. These results indicate the mean performance across all models on their respective local datasets.

These results reveal that regularisation without immediate replacement of the local classifier's parameters has a positive impact on the model performance, especially when the number of features per client is low. This is reflected in DivEn's superior performance compared to all other methods. However, this superiority either plateaus or is surpassed by the DivEn-mix strategy as the number of features increases. This indicates that when the number of features per client grows, directly replacing the parameters of the local classifier with those of the global classifier (obtained after aggregation) improves performance.

\begin{table}[h!]
\centering
\caption{Performance Comparison of Methods Across Executions on Obesity and Lifestyle Datasets }
\label{tab:side_by_side_tables}
\resizebox{0.50\textwidth}{!}{%
\begin{tabular}{c@{\hspace{5pt}}c}
\begin{tabular}{|l|c|c|}
\hline
\textbf{Method} & \multicolumn{2}{|c|}{\textbf{Obesity Dataset Acc. (\%)}} \\
\hline
                & \textbf{8 Features} & \textbf{12 Features} \\
\hline
Single         & 60.15 & 70.28 \\
Auto\_encoder  & 40.47 & 47.37 \\
Align\_corr    & 62.17 & 71.38 \\
Class\_agg     & 59.08 & 71.22 \\
DivEn          & \textbf{64.81} & 72.10 \\
DivEn-mix      & 63.73 & \textbf{73.37} \\
\hline
\end{tabular}
&
\begin{tabular}{|c|c|}
\hline
\multicolumn{2}{|c|}{\textbf{Lifestyle Dataset MAE}} \\
\hline
\textbf{8 Features} & \textbf{12 Features} \\
\hline
5.77 & 4.87 \\
6.54 & 7.21 \\
4.07 & 3.51 \\
2.53 & 2.28 \\
\textbf{2.27} & \textbf{1.89} \\
2.56 & 1.98 \\
\hline
\end{tabular}
\end{tabular}
}
\end{table}

Building upon the observation discussed above, the development of DivEn-c enabled the maximisation of aggregation opportunities for client models exhibiting high feature similarity. The following Fig.~\ref{fig:divEn_c} and Tab.~\ref{tab:comparison_diven_c} demonstrate DivEn-c's consistent superiority across an additional series of executions.

\begin{figure}[h]
\centering
\vspace{0in}
    \includegraphics[width=9cm]{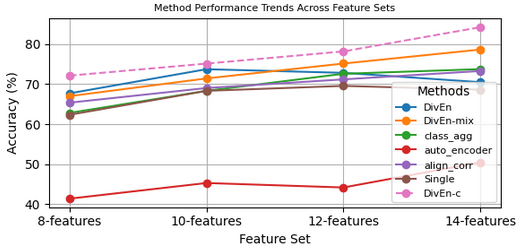}
    \vspace{0.1in}
    \caption{Method comparison across different feature set distribution.}
    \label{fig:divEn_c}
   \vspace{-0.1in}
\end{figure}

\begin{table}[h!]
\centering
\caption{Accuracy (\%) of Methods Across Feature Set Sizes}
\resizebox{0.50\textwidth}{!}{%
\begin{tabular}{|l|c|c|c|c|}
\hline
\textbf{Method} & \textbf{8 Features} & \textbf{10 Features} & \textbf{12 Features} & \textbf{14 Features} \\
\hline
Single         & 62.33 & 68.4 & 69.53 & 68.60 \\
\hline
Auto\_encoder  & 41.40 & 45.34 & 44.19 & 50.47 \\
\hline
Align\_corr    & 65.35 & 64.42 & 71.16 & 73.26 \\
\hline
Class\_agg     & 62.79 & 69.07 & 72.56 & 73.72 \\
\hline
DivEn          & 67.67 & 73.72 & 72.79 & 70.47 \\
\hline
DivEn-mix      & 66.98 & 71.39 & 75.12 & 78.60 \\
\hline
DivEn-c        & \textbf{72.09} & \textbf{75.11} & \textbf{78.14} & \textbf{84.19} \\
\hline
\end{tabular}
}
\label{tab:comparison_diven_c}
\end{table}

\begin{figure}[h]
\centering
\vspace{0in}
    \includegraphics[width=9cm]{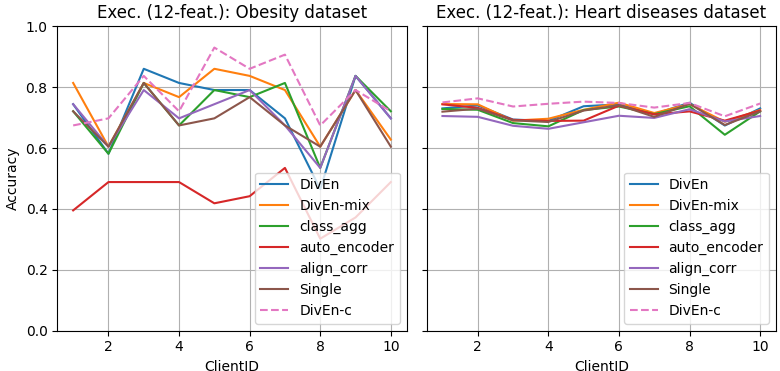}
    \vspace{0.1in}
    \caption{Client by client comparison on Obesity and Heart diseases datasets}
    \label{fig:divEn_c_clientByclient}
   \vspace{-0.1in}
\end{figure}

Fig.~\ref{fig:divEn_c_clientByclient} presents a comparison of individual client performance on the Obesity and Heart Diseases datasets. The figure indicates an improvement in accuracy for a subset of the 10 clients utilised in our experiments.

\begin{table}[h]
\centering
\caption{Performance Comparison on Obesity and Heart Disease Datasets (12 Selected Features)}
\begin{tabular}{|l|c|c|}
\hline
\textbf{Method}       & \textbf{Obesity Dataset} & \textbf{Heart Disease Dataset} \\
\hline
Single                & 69.5      & 71.3  \\ 
\hline
AutoEncoder           & 44.2     & 71.3   \\
\hline
Align-Corr            & 71.2     & 69.6   \\
\hline
Class-Agg             & 72.6     & 70.8   \\
\hline
DivEn                 & 72.8     & 72.0   \\
\hline
DivEn-Mix             & 75.1     & 72.2   \\
\hline
DivEn-C               & \textbf{78.1}  & \textbf{74.3} \\
\hline
\end{tabular}
\label{tab:comparison_plain}
\end{table}

\subsection{Domain Adaption}
This section presents experimental results on two datasets: Digits-Five \cite{digit5} dataset utilised for benchmarking; and Pneumonia Chest X-rays dataset \cite{xrays} employed as an e-health dataset. The experiments are carried out using the Flower FL framework \cite{flowerfl} within the Google Colab environment for implementation. We evaluate four algorithms: FADA, FedKA, FedDAFL (without accuracy boost), and FedDAFL* (with accuracy boost).

\subsubsection{Digits-Five Dataset}
\begin{figure}[h]
\centering
\vspace{-0.0in}
    \includegraphics[width=9.25cm]{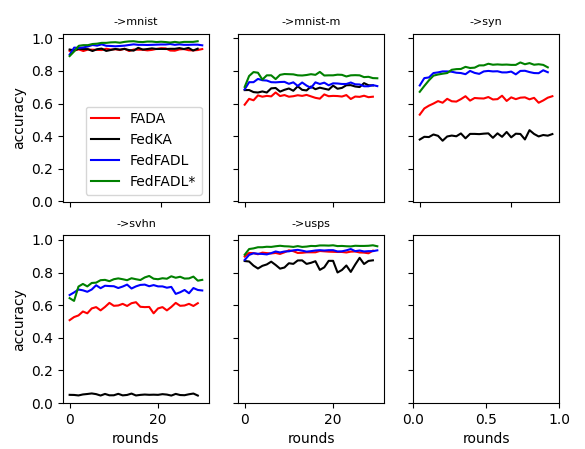}
    \vspace{-0.1in}
    \caption{Accuracy on Digits-Five dataset. }
    \label{fig:resultdigit5}
   \vspace{-0.05in}
\end{figure}

The Digits-Five dataset encompasses five of the most prominent digit datasets, each contributing a unique style to the collection: MNIST (mt) of 55,000 samples, MNIST-M (mm) of 55,000 samples, Synthetic Digits (syn) of 25,000 samples, SVHN (sv) of 73,257 samples, and USPS (up) of 7,438 samples. Each dataset presents a distinct visual representation of digits from 0 to 9. In our experiments, we employed 5,000 samples per dataset (per node) for training and 1,000 samples for testing. In Figure~\ref{fig:resultdigit5}, the relative accuracy plots of our approach, comprising two models, are compared to FADA and FedKA's accuracy plots across three scenarios on the Digits-Five dataset. The results demonstrate that the proposed models FedDAFL and FedDAFL* outperform the two existing algorithms for all rounds. 

Similarly, Table \ref{table:digit5results} provides a detailed account of the results obtained from our approach, presenting a side-by-side comparison with FADA and FedKA. The reported accuracy values represent the highest achieved during the execution. To conform to FADA and FedKA's execution conditions, all datasets, with the exception of the target, are considered as the source. This meticulous consideration ensures a fair comparison and aligns with the established standards of execution conditions.

\begin{table}[h]
\vspace{-0.05in}
    \caption{Results on Digits-Five dataset}
            \vspace{-0.01in}
            \centering
            \label{table:digit5results}
            \vspace{-0.05in}
            \begin{tabular}{ |c|c|c|c|c|c|} 
             \hline
                {\bf Methods} & {\bf $\rightarrow$ mt} & {\bf $\rightarrow$ mm} &   {\bf $\rightarrow$ syn} & {\bf $\rightarrow$ sv} & {\bf $\rightarrow$ up} \\
            \hline
                FADA        &  93.9   & 66.7  &  64.4   & 61.8 & 93.5\\ 
                FedKA        &  94.1   & 72.4  &  45.5   & 16.1 & 89.7\\ 
                FedDAFL        &  97.1   & 75.0  &  80.0   & 72.6 & 94.4\\
                FedDAFL*       &  98.2   & 79.2  &  85.9   & 77.7 & 96.7\\
            \hline
            \end{tabular}
        \vspace{-0.15in}
\end{table}

Table \ref{table:multi_destination_res} presents the outcomes of the Multi-Teacher Multi-Learner execution with the acronyms described in the Digits-Five dataset. Unlike FADA, which necessitated training the source domains individually with each destination domain, FedDAFL produced these results in a single execution. The results exhibit a notable difference, showcasing the effectiveness of FedDAFL compared to FADA and FedKA.

\begin{table}[h]
\centering
\begin{scriptsize}
\vspace{-0.05in}
\caption{Multi-Teacher Multi-Learner Execution }
\label{table:multi_destination_res}
\vspace{-0.05in}
\begin{tabular}{ |c|c|c|c|c|c|c|c|c| } 
\hline
{\bf Teachers $\rightarrow$} &\multicolumn{2}{|c|}{\bf mm, sv, up}&\multicolumn{2}{|c|}{\bf mt, syn, up} &\multicolumn{2}{|c|}{\bf mt, mm, up} &\multicolumn{2}{|c|}{\bf mt, sv, syn}\\
 \hline
{\bf Learners $\rightarrow$} & {\bf mt} & {\bf syn} & {\bf mm} & {\bf sv} & {\bf syn} & {\bf sv} & {\bf mm} & {\bf up} \\
\hline\hline
 FADA & 93.7  &  66.0 & 56.0 &51.1  &  39.5 & 27.9 &64.2 & 89.1\\
 FedKA & 92.6  &  46.0 & 72.0 &14.1  &  57.3 & 14.6 &64.3 & 89.4\\
 FedDAFL & 96.2  &  81.5 & 78.7 &76.3  &  63.2 & 57.8 &74.2 & 94.7\\
 FedDAFL* & 96.7  &  84.5 & 78.6 &78.6  &  70.2 & 65.5 &78.6 & 94.9\\
\hline
\end{tabular}
\end{scriptsize}
\vspace{-0.1in}
\end{table}

While Table~\ref{table:multi_destination_res} focuses on the accuracy comparisons during Multi-Teacher Multi-Learner (MT-ML) execution, Table~\ref{table:execution_time} delves into the execution time achieved by Multi-Teacher Single-Learner (MT-SL) and MT-ML for the three models. The reported results in Table~\ref{table:execution_time} underscore the significant performance advantage of both FedDAFL and FedDAFL* models over FADA, but with FedKA emerging as the top-performing model in terms of execution time. Note that, although FedDAFL* exhibits superior accuracy compared to FedDAFL due to pseudo-labelling, it lags behind FedDAFL in terms of execution time because of the additional time required for pseudo-labelling.
\begin{table}[h]
\centering
\vspace{-0.1in}
\caption{Average Execution Time}
\label{table:execution_time}
\vspace{-0.05in}
\begin{tabular}{ |c|c|c| } 
\hline
{\bf Method} & {\bf MT-SL} & {\bf MT-ML}\\
\hline
 FADA & 6590.8  &  9512.4\\
 FedKA & 328.2  &  511.0 \\
 FedDAFL & 473.3  &  895.4 \\
 FedDAFL* & 1361.2  &  1821.1 \\
\hline
\end{tabular}
\vspace{-0.1in}
\end{table}

\vspace{-0.05in}
\subsubsection{Chest X-rays Dataset}
The Chest X-rays (Pneumonia) dataset  consists of 5,863 X-ray images categorized into Pneumonia/Normal. In our experiments, we used a subset of 1,076 training images and 267 test images from 1,000 different patients. We conducted two setss of experiments on three clients in both IID and non-IID settings. In the IID setting, each labelled client has an almost balanced number of samples per class, while the non-IID setting explores entirely unbalanced scenarios where each labelled client has data of only one class.

\begin{table}[h]
\centering
\begin{scriptsize}
\vspace{-0.05in}
\caption{Experimental setting}
\label{table:experimentsconfig}
\vspace{-0.05in}
\begin{tabular}{ |p{0.005\textwidth}|p{0.01\textwidth}|c|c|c|c|c|c|c|c|c| } 
\hline
 &&\multicolumn{3}{|c|}{\bf Setup 1}&\multicolumn{3}{|c|}{\bf Setup 2} &\multicolumn{3}{|c|}{\bf Setup 3} \\
 \hline
                    & {\bf Cl.}  & {\bf Total} & {\bf C1} & {\bf C2} & {\bf Total} & {\bf C1} & {\bf C2} &  {\bf Total} & {\bf C1} & {\bf C2}\\
\hline
\multirow{1}{2.0em}{\rotatebox{90}{\bf IID}} & 1  & 292  & 146  & 146 &100 &38 &62 &37 &16 &21\\ 
                            & 2   & 311 & 151  & 160 &299 &166 &133 &84 &31 &53  \\ 
                            & 3 & 504 &  - & - &708&-&-&953&-&-\\ 
\hline\hline
\multirow{1}{2.0em}{\rotatebox{90}{\bf Non-IID}} & 1  & 308  & 308  & 0 &207 &207 &0 &52 &52 &0\\ 
                            & 2   & 297 & 0  & 297 &194 &0 &194 &71 &0 &71  \\ 
                            & 3 & 504 &  - & - &708&-&-&953&-&-\\ 
\hline\hline
\end{tabular}
\end{scriptsize}
\vspace{-0.08in}
\end{table}

\begin{figure}[h]
\centering
\vspace{-0.2in}
    \includegraphics[width=9cm]{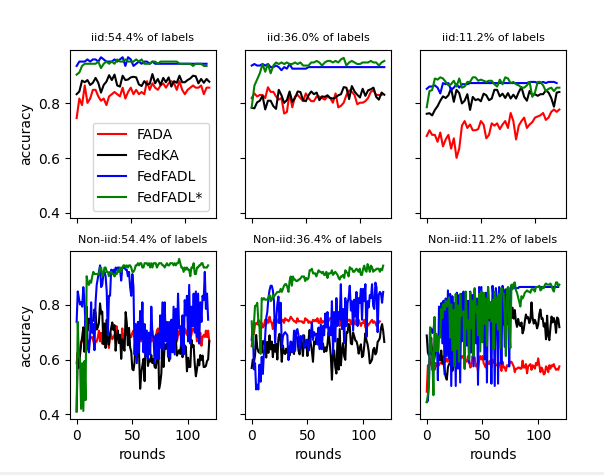}
    \vspace{-0.25in}
    \caption{Accuracy in Percentage}
    \label{fig:xrays_curves}
\end{figure}

While Table ~\ref{table:experimentsconfig} outlines the experimental settings for comparing FADA, FedKA, and our proposed model, Table \ref{table:results} and Fig.~\ref{fig:xrays_curves} present the results of our experiments.

\vspace{0.13in}

\begin{table}[h]
\centering
\vspace{-0.05in}
\caption{Results on X-rays dataset}
\label{table:results}
\vspace{-0.05in}
\begin{tabular}{ |c|c|c|c|c| } 
\hline
 {\bf Case} & {\bf Method} & {\bf Setup 1} & {\bf Setup 2} & {\bf Setup 3} \\
\hline
\multirow{1}{2.5em}{\centering\rotatebox{90}{IID}} & FADA  & 88.0  & 85.8  & 77.7\\ 
                            & FedKA  & 86.0  & 89.1  & 88.8\\ 
                            & FedDAFL   & 96.8 & 94.3  & 87.8   \\ 
                            & FedDAFL* & 96.0 &  96.6 & 89.4 \\ 
\hline\hline
\multirow{1}{2.5em}{\centering\rotatebox{90}{Non-IID}} & FADA  & 73.8  & 76.8  & 61.3\\ 
                            & FedKA   & 80.5 & 73.0  & 81.6  \\ 
                            & FedDAFL   & 94.4 & 88.1  & 86.9   \\ 
                            & FedDAFL* & 96.0 &  94.9 & 88.2 \\                            
\hline
\end{tabular}
\vspace{-0.15in}
\end{table}

Figure~\ref{fig:xrays_curves} illustrates that the accuracy of our FedDAFL model consistently surpasses FADA and FedKA for the three different scenarios and across different rounds. Moreover, Table \ref{table:results} details the performance comparison between our model and the othet two algorithms, demonstrating the superiority of our model for both IID and non-IID cases. This underscores the relative performance of our model compared to FADA and FedKA in the experiments with Chest X-rays dataset.

\vspace{-0.05in}
\section{Performance Justification} \label{sec:5}
\subsection{DivEn-C}
The superior performance of DivEn-C can be attributed to three key strategies, which collectively enhance generalisation and personalisation across clients.
\subsubsection{Learning with Optimised Hyperparameters}
It is well-established that training models with optimal hyperparameters improves predictive performance~\cite{automl0,hpoptim}. Building on this principle, DivEn-C incorporates hyperparameter optimization—performed prior to the federated learning process—for each client or client cluster. This initialisation ensures that local models are primed with well-tuned settings before collaborative training begins.

Formally, we expect that for a client $i$, the trained model using an optimal hyperparameter vector $\mathbf{h}^*$ satisfies:
\[
\text{performance}(\mathcal{M}_i, X_i, \mathbf{h}^*) > \text{performance}(\mathcal{M}_i, X_i, \mathbf{h}), \quad \forall \mathbf{h} \neq \mathbf{h}^*
\]
where $\mathcal{M}_i$ is the client's model, $X_i$ is its dataset, and $\mathbf{h}$ denotes any arbitrary hyperparameter configuration.
\subsubsection{Effect of Commonality Sharing}
The personalisation mechanism in DivEn is anchored in the construction of a weighted global classifier $\mathcal{C}_{G_i}$ for each client $i$, derived from latent-space similarity across all clients. This aggregation step shares commonality across peers while preserving personalised relevance, effectively transferring structured knowledge from similar learners to the target client. Subsequently, the local training loss integrates a regularisation term:
\[
\mathcal{L}_i = \text{CE}(y_i, \hat{y}_i) + \lambda \sum_l \left\| \theta_{\mathcal{C}_i, l} - \theta_{\mathcal{C}_{G_i}, l} \right\|_2^2
\]
which ensures that the personalized classifier $\mathcal{C}_i$ remains closely aligned with $\mathcal{C}_{G_i}$ while retaining local adaptability.
Together, the similarity-weighted aggregation and the customised regularised loss promote a balance between global generalisation and local specialisation. This hybrid learning strategy enables each model $\mathcal{M}_i$ to benefit from distributed experiences while remaining tailored to its own data—contributing to its strong personalized performance.

\subsubsection{Effect of Client Clustering}
The feature-based client clustering strategy aims to group clients with high feature-space similarity, thereby enabling the aggregation of their encoders and classifiers. This approach leverages all samples within each cluster, increasing representational diversity and stabilising learning dynamics.

From a convergence perspective, this mechanism aligns with the principles of federated averaging (FedAvg), which establishes that the expected gap between the global objective and local updates diminishes over communication rounds when (1) The learning rate is properly bounded, (2) The loss functions are convex and Lipschitz continuous, and (3) Each client contributes proportionally to its dataset size.

As shown,  the convergence rate under non-IID data and multiple clients is bounded as:
\[
\mathbb{E}\left[F(\bar{\mathbf{w}}_T)\right] - F(\mathbf{w}^*) \leq \mathcal{O}\left(\frac{1}{\sqrt{T}}\right),
\]
where $\bar{\mathbf{w}}_T$ is the averaged model after $T$ rounds and $F(\mathbf{w}^*)$ is the optimal centralized loss.

By clustering clients with similar features and aggregating them proportionally based on dataset sizes, DivEn-C benefits from FedAvg's convergence behaviour while reducing intra-cluster heterogeneity, which accelerates convergence and improves final model performance.

\subsection{Domain Adaptation}
The superior performance of FedDAFL is attributed to two foundational mechanisms:
\subsubsection{Construction of a Robust Global Encoder}
FedDAFL enhances its encoder through diverse training signals received from both labeled and unlabeled clients:
\begin{itemize}
    \item Labeled Clients: Optimize the supervised classification loss:
    \[
    \mathcal{L}_m = -\sum_{c=1}^{C} y_c \log\left( \text{softmax}(M(E(x)))_c \right)
    \]
    \item Unlabeled Clients: Optimize the self-supervised pretext loss using rotation prediction:
    \[
    \mathcal{L}_p = -\sum_{r=1}^{R} \tilde{y}_r \log\left( \text{softmax}(P(E(\text{Rot}(u))))_r \right)
    \]
\end{itemize}
Each client minimises its respective loss, and the server aggregates encoders each round:
\[
\theta_e(t) = \sum_{k=1}^{N} \theta_{e_k}(t).
\]
This combines semantic separability from $\mathcal{L}_m$ with domain-invariant structure from $\mathcal{L}_p$, yielding a task-aware, cross-domain encoder.

\subsubsection{Learning with Soft Labels}
After encoder pretraining, a global $E{+}M$ is distributed to all clients.
\begin{itemize}
    \item Unlabeled clients adopt pseudo-labelling, accepting predictions with confidence $\ge\tau$ and optimising the filtered cross-entropy in Eq.~\eqref{e:l2}.
\end{itemize}
Thus, FedDAFL enables distributed learning in non-IID, label-scarce settings via encoder fusion and adaptive supervision.

\vspace{-0.05in}
\section{Conclusion (\textbf{FedFusion})} \label{sec:6}
\vspace{-0.05in}
This study introduced \textbf{FedFusion}, a federated transfer learning framework designed to address the complex challenges of non-IID data distributions in federated environments through the integration of domain adaptation and frugal labelling strategies. The proposed framework incorporates DivEn-c, a diversity-aware encoding and clustered aggregation approach, and leverages personalised encoder–classifier architectures to accommodate heterogeneous feature spaces and varying degrees of label availability across clients. Our evaluations across tabular and image-based healthcare datasets demonstrated that \textbf{FedFusion} consistently outperforms baseline models, particularly in scenarios characterised by missing features, domain shifts, and label scarcity.

The design of \textbf{FedFusion} provides a generalisable and scalable solution to the dual challenge of data heterogeneity and limited supervision in privacy-preserving, decentralised learning environments. Notably, the framework maintains flexibility to support both cross-silo and cross-device learning paradigms, positioning it as a viable tool for real-world applications where data is fragmented and labelling is costly.

Future research will explore the extension of \textbf{FedFusion} to accommodate dynamic client participation and asynchronous communication protocols, as well as its integration with explainable AI mechanisms to enhance model interpretability in high-stakes domains such as healthcare. Additionally, the development of benchmarking suites for evaluating FTL systems under diverse non-IID conditions will be a critical step towards standardising performance assessment and advancing the robustness of federated learning methodologies.

\end{document}